\documentclass{article}

\usepackage[square]{natbib}

\usepackage[preprint]{neurips_2023}

\usepackage[utf8]{inputenc} 
\usepackage[T1]{fontenc}    
\usepackage{hyperref}       
\usepackage{url}            
\usepackage{booktabs}       
\usepackage{amsfonts}       
\usepackage{nicefrac}       
\usepackage{microtype}      
\usepackage{xcolor}         
\usepackage{graphicx}
\usepackage{listings}
\usepackage{framed}
\usepackage{subcaption}
\usepackage{threeparttable}
\usepackage{amsmath}

\title{Augmented Large Language Models with\\Parametric Knowledge Guiding}

\author{Ziyang Luo$^{1}$\thanks{Work done during the internship at Microsoft.}, Can Xu$^{2}$, Pu Zhao$^{2}$, Xiubo Geng$^{2}$, Chongyang Tao$^{2}$,\\\textbf{Jing Ma}$^{1}$, \textbf{Qingwei Lin}$^{2}$, \textbf{Daxin Jiang}$^{2}$\thanks{ Corresponding author}\\
  $^1$  {\normalsize Hong Kong Baptist University, Hong Kong SAR, China} \\
  $^2$ {\normalsize Microsoft Corporation} \\
    \texttt{\normalsize cszyluo@comp.hkbu.edu.hk, majing@hkbu.edu.hk}\\
  \texttt{\normalsize \{caxu,pu.zhao,xigeng,chongyang.tao,qlin,djiang\}@microsoft.com}}

\begin{document}

\maketitle

\begin{abstract}

Large Language Models (LLMs) have significantly advanced natural language processing (NLP) with their impressive language understanding and generation capabilities. However, their performance may be suboptimal for domain-specific tasks that require specialized knowledge due to limited exposure to the related data. Additionally, the lack of transparency of most state-of-the-art (SOTA) LLMs, which can only be accessed via APIs, impedes further fine-tuning with domain custom data. Moreover, providing private data to the LLMs' owner leads to data privacy problems. To address these challenges, we propose the novel \textbf{Parametric Knowledge Guiding (PKG)} framework, which equips LLMs with a knowledge-guiding module to access relevant knowledge without altering the LLMs' parameters. Our PKG is based on open-source "white-box" language models, allowing offline memory of any knowledge that LLMs require. We demonstrate that our PKG framework can enhance the performance of "black-box" LLMs on a range of domain knowledge-intensive tasks that require factual ($+7.9\%$), tabular ($+11.9\%$), medical ($+3.0\%$), and multimodal ($+8.1\%$) knowledge.

\end{abstract}

\section{Introduction}

Large Language Models (LLMs) such as GPT-family~\citep{gpt4} have exhibited impressive proficiency across a diverse range of NLP tasks. These models are typically trained on extensive data from the internet, thereby enabling them to assimilate an immense amount of implicit world knowledge into their parameters. As a result, LLMs have emerged as versatile tools that find numerous applications in both research and industry. For instance, they can be used for machine translation~\citep{gpt4mt}, document summarization~\citep{gpt4sum}, and recommendation systems~\citep{gpt4rec}. With their exceptional language understanding and generation capabilities, LLMs have opened up new opportunities for diverse industrial applications, such as the recently launched New Bing~\citep{newbing} and ChatGPT Plugins~\citep{plugins}.

\begin{figure}
    \centering
    \includegraphics[height=7cm]{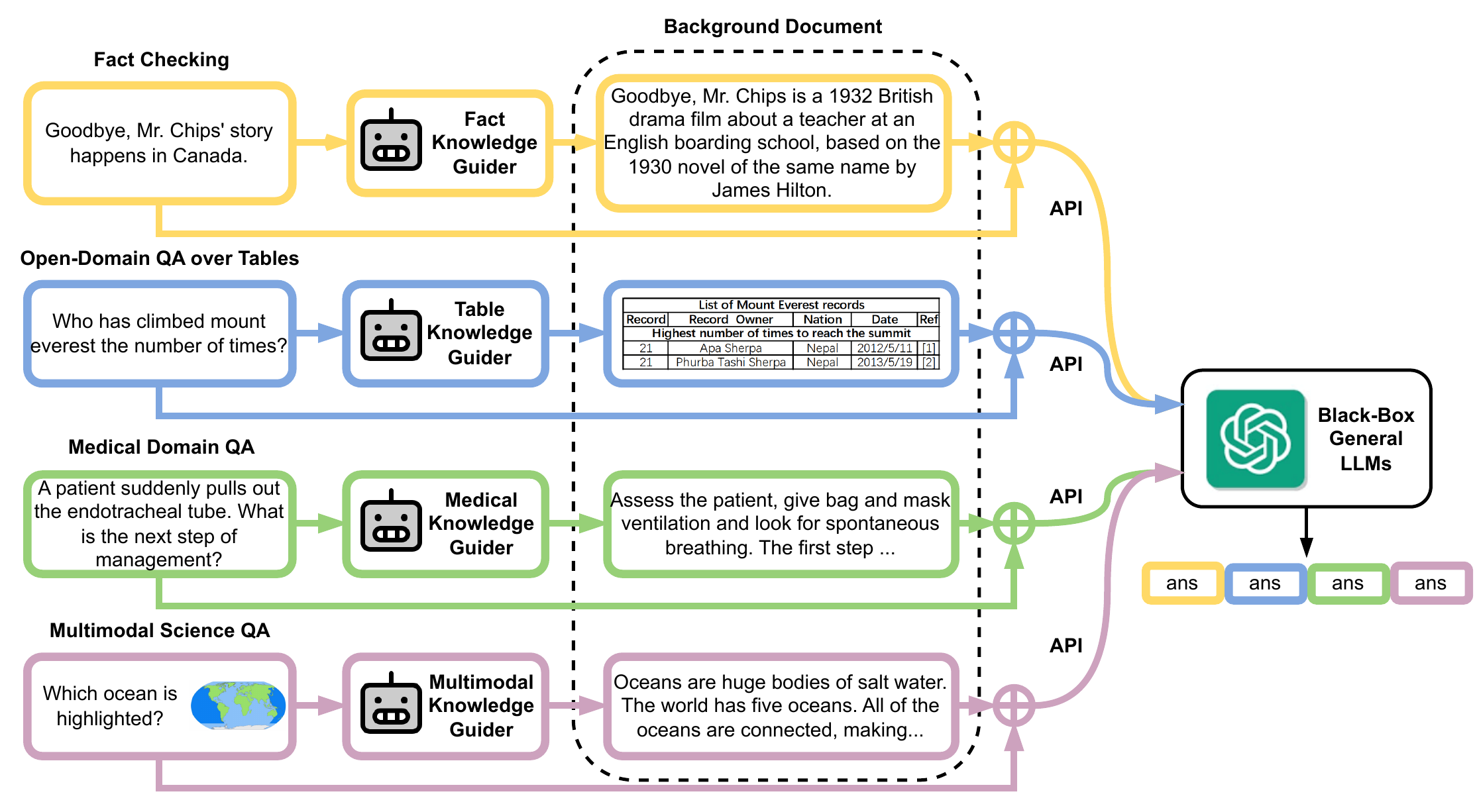}
    \caption{A brief introduction of our parametric knowledge guiding framework (PKG) for augmenting "black box" LLMs on domain-specific tasks.}
    \label{fig:intro}
\end{figure}

Despite their impressive performance across various general tasks, LLMs may face challenges when applied to domain-specific tasks~\citep{chatgpt4law,gpt4med,gpt4chem} due to their limited exposure to relevant knowledge and vocabulary. Although LLMs acquire implicit world knowledge during pre-training, such knowledge may be insufficient or inappropriate for specific tasks, resulting in less effective performance. Furthermore, many state-of-the-art LLMs are considered "black-box" models, accessible only through APIs. This lack of transparency presents significant challenges and high costs for most researchers and companies seeking to fine-tune these models for their specific use cases or domains. Moreover, users who can afford to fine-tune must provide their private data to the LLMs' owner, thereby exposing it to potential risks such as misuse, breaches, or other security threats~\citep{gptban}. These limitations hinder the adaptability of LLMs to diverse scenarios and domains.

A common approach to enhance LLMs is to leverage retrieval-based methods that access domain-specific knowledge from external sources~\citep{LlamaIndex,replug,DBLP:journals/corr/abs-2302-12813}. While these methods have shown promise, they face several challenges. First, they heavily rely on modern dual-stream dense retrieval models~\citep{dpr} which suffer from shallow interaction between the query and candidate documents. Second, most dense retrieval models are based on small-scale pre-trained models such as BERT~\citep{bert} and therefore cannot take advantage of the world knowledge of large-scale pre-trained models. Third, retrieval models may struggle with complex knowledge that requires the integration of information from multiple sources or modalities.

In this work, we propose the \textbf{Parametric Knowledge Guiding (PKG)} framework, which enables LLMs to access relevant information without modifying their parameters, by incorporating a trainable background knowledge generation module, as illustrated in Figure~\ref{fig:intro}. Unlike retrieval-based methods, our PKG module utilizes open-source and free-to-use "white-box" language models, LLaMa-7B~\citep{llama}, which encode implicit world knowledge from large-scale pre-training. The framework consists of two steps. First, we align the PKG module with the specific task or domain knowledge via instruction fine-tuning~\citep{instructGPT} to capture the necessary expertise. Second, for a given input, the PKG module generates the related knowledge, fed as extra context to the background-augmented prompting for LLMs. By supplying the necessary knowledge, our framework can enhance the performance of LLMs on domain-specific tasks.

Our experiments demonstrate that the proposed PKG framework enhances the performance of "black-box" LLMs on various downstream tasks which require domain-specific background knowledge, including factual knowledge (FM2~\citep{fm2}, $+7.9\%$), tabular knowledge (NQ-Table~\citep{nq_table}, $+11.9\%$), medical knowledge (MedMC-QA~\citep{medqa}, $+3.0\%$), and multimodal knowledge (ScienceQA~\citep{sciqa}, $+8.1\%$). 

We summarize our contributions as follows:
\begin{itemize}
    \item We propose a novel \textbf{Parametric Knowledge Guiding (PKG)} framework that integrates a background knowledge generation module to enhance the performance of LLMs on domain-specific tasks.
    \item We introduce a knowledge-guiding process by first aligning the parametric modules with specific tasks or domain knowledge and then generating related knowledge as the extra context in the background-augmented prompting.
    \item We conduct extensive experiments on various downstream tasks to evaluate the effectiveness of our proposed PKG framework. The experiments demonstrate that our PKG framework can improve the capability of LLMs on domain knowledge-intensive tasks.
\end{itemize}

\section{Related Work}

\paragraph{Large Language Models.}
LLMs, such as GPT3~\citep{gpt3}, Codex~\citep{codex}, PaLM~\citep{palm}, and GPT4~\citep{gpt4}, have gained widespread attention due to their remarkable language understanding and generation capabilities~\citep{cot_first,DBLP:journals/corr/abs-2210-03057}. However, their performance can be limited when it comes to domain-specific tasks, where they may lack exposure to specialized knowledge and vocabulary~\citep{chatgpt4law,gpt4med,DBLP:journals/corr/abs-2303-01067}. Moreover, while some SOTA LLMs such as InstructGPT3.5 and ChatGPT~\citep{instructGPT} exist, they are available only as "black box" APIs due to commercial considerations. This limits researchers and developers with limited resources, who may not be able to access or modify the models' parameters. While open-source LLMs such as OPT-175B~\citep{opt} and BLOOM-176B~\citep{bloom} are available, they lag significantly behind SOTA LLMs on most tasks. Additionally, running and fine-tuning these open LLMs locally requires significant computational resources.

\paragraph{Augmented Large Language Models.}
ALLMs are a recent popular topic in NLP that aim to enhance the context processing ability of LLMs by incorporating external modules~\citep{ALMs,visualgpt,huggingGPT,Chameleon,audiogpt}. One approach to achieving this goal is through the use of retrieval-augmented large language models (RLLMs)\citep{ralp,atlas,incontext,replug}. RLLMs leverage external knowledge by retrieving relevant documents or passages from knowledge sources using retrieval-based methods such as BM25\citep{bm25} and DPR~\citep{dpr}. These retrieved passages are then used as additional contexts to improve the LLMs' performance on the task at hand. Although RLLMs have shown promise in enhancing LLMs' performance, they have certain limitations. For instance, they rely heavily on the dual-stream dense retriever, which leads to shallow interaction between the query and the candidate information. Furthermore, they may struggle with complex queries that require integrating information from multiple sources or modalities.

\paragraph{Instruction Fine-Tuning.}
IFT is a technique in NLP that aims to align language models with specific user intents~\citep{instructGPT}. While many LLMs are trained on large datasets of internet data to predict the next word, they may not be tailored to the specific language tasks that users require, meaning that these models are not inherently aligned with their users' needs. Recent research~\citep{flan,t0,zeroprompt,DBLP:conf/emnlp/XieW0ZSYWZYWZWL22} has highlighted the potential of IFT as a key technique for improving the usability of LLMs. Our proposed approach, PKG, follows the same principle of aligning the basic module with task-specific knowledge to enhance its performance.

\section{Parametric Knowledge Guiding for LLMs}

In this section, we present our \textbf{PKG} framework to guide the reasoning process of LLMs on domain-specific tasks. These tasks differ from general tasks such as document summarization due to their reliance on specific background knowledge. However, this knowledge may be absent or incomplete in the LLMs' training data. Furthermore, continuous pre-training of LLMs with domain knowledge poses several challenges: (1) limited transparency of accessing current SOTA LLMs solely through APIs, (2) the potentially high fine-tuning cost associated with APIs usage, and (3) concerns regarding data privacy when providing private data to LLMs' owners. To tackle these issues, we adhere to the \textit{generate-then-read} paradigm~\citep{genread} and leverage an offline PKG module to generate relevant background knowledge. Our method is first formulated in §~\ref{subsec:formula}. Next, we describe the background knowledge alignment of our PKG modules in §~\ref{subsec:align}. Finally, we introduce background-augmented prompting for LLMs in §~\ref{subsec:promting}.


\subsection{Formulation}~\label{subsec:formula}
Given a question/input $\mathcal{Q}$ associated with some contexts, LLMs take the input and generate a response by maximum a posteriori estimation (MAP):
\begin{equation}
\hat{\mathcal{A}}:=\text{argmax}_{\mathcal{A}}P(\mathcal{A}|\mathcal{Q},\mathcal{M}^{\textit{LLM}}),
\end{equation}
where $\mathcal{M}^{\textit{LLM}}$ represents the parameters of the LLMs. However, for tasks that require background knowledge beyond what is contained in the input, such as knowledge-intensive tasks, relying solely on LLMs may not be effective. This is because there may be a significant amount of additional domain-specific knowledge that remains unexploited.

To improve performance, we first introduce an auxiliary PKG module $\mathcal{M}^{\textit{PKG}}$ to align specific background knowledge (§\ref{subsec:align}). Next, we estimate the input-related background knowledge $\mathcal{K}$ using MAP estimation:
 \begin{equation}
\hat{\mathcal{K}}:=\text{argmax}_{\mathcal{K}}P(\mathcal{K}|\mathcal{Q},\mathcal{M}^{\textit{PKG}}).
\end{equation}
Finally, the background knowledge $\mathcal{K}$ enriches the input by incorporating background-augmented prompting for LLMs (§~\ref{subsec:promting}) in the form:
\begin{equation}
P(\mathcal{A}|\mathcal{Q}):=P(\mathcal{A}|\mathcal{K},\mathcal{Q},\mathcal{M}^{\textit{LLM}})P(\mathcal{K}|\mathcal{Q},\mathcal{M}^{\textit{PKG}}).
\end{equation}


\begin{figure}
    \centering
    \includegraphics[height=3cm]{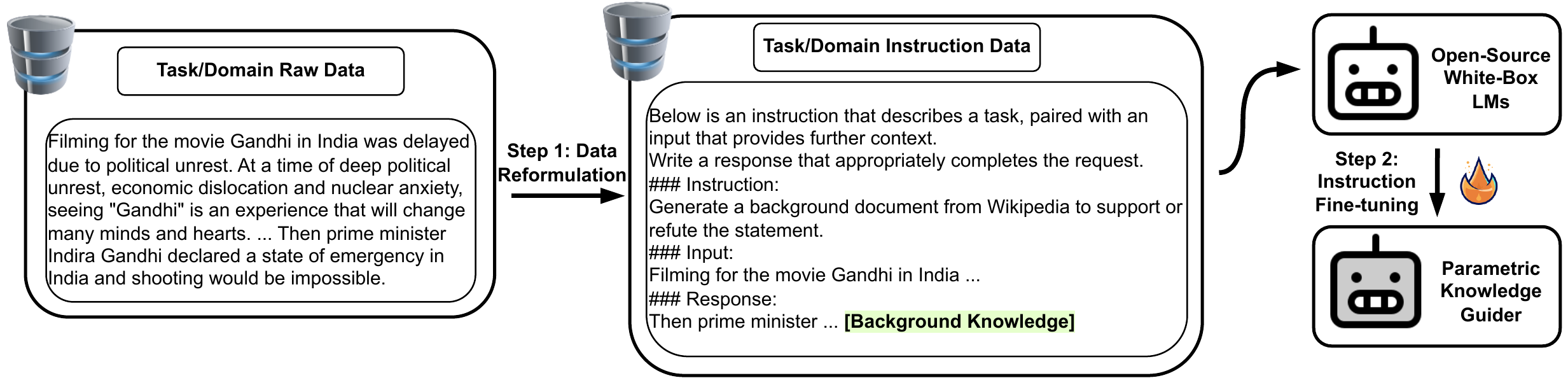}
    \caption{The knowledge alignment example of the PKG module on the fact-checking task (FM2). The passage behind the "Response" is the background knowledge of the "Input".}
    \label{fig:method}
\end{figure}

\subsection{Background Knowledge Alignment}\label{subsec:align}

Given a target task or domain, our PKG framework utilizes an open-source language model to align with relevant knowledge. Figure~\ref{fig:method} presents an example of the fact-checking task. This process is divided into two steps. First, we collect raw data about the target task/domain, which serves as our background knowledge. Second, we transform the data into a set of (instruction, input, output) triples. The instruction serves as a prompt for the input and guides the module to align with the expected knowledge.

Next, this set of triples is adopted to tune our basic PKG module with instruction fine-tuning~\citep{instructGPT}, which optimizes its ability to provide relevant and effective background knowledge to the LLMs. This two-step process can be completed fully offline, without requiring us to provide our private data to tune the LLMs. Once aligned with the task background knowledge, the PKG module learns to generate domain-specific knowledge to assist the LLMs during runtime.

The instruction data format of the fact-checking task is:

\texttt{\scriptsize
Below is an instruction that describes a task, paired with an input that provides further context.\\
Write a response that appropriately completes the request.\\
\#\#\# Instruction:\\
<instruction>\\
\#\#\# Input:\\
<input sentence>\\
\#\#\# Response:\\
<background>
}

The \texttt{<input sentence>} is a sentence within the specified task. The \texttt{<background>} is the background knowledge that the model generates based on the given \texttt{<instruction>} and \texttt{<input sentence>}. The basic PKG module is trained in a standard supervised way with an auto-regressive manner, where the model generates the \texttt{<background>} given the previous context. More instruction data formats for different tasks are presented in Appendix F. 

\subsection{Background-Augmented Prompting}\label{subsec:promting}

Instead of directly requesting the LLMs to generate the answer or response for the input question or sentence via APIs, we first instruct the PKG module to generate the background knowledge. In the second step, we utilize the generated background in combination with the input question to derive the final answer from the LLMs. This is similar to the "zero-shot" open-domain question-answering setting that has been widely explored in prior research~\citep{gpt3,DBLP:journals/corr/abs-2203-05115,genread}. The background-augmented prompt of the fact-checking task is:

\texttt{\scriptsize
<background>\\
Claim: <input sentence>\\
Is the claim true or false?
}

Finally, the augmented prompt is fed into the LLMs to generate an answer. More prompts for different tasks are presented in Appendix G.

\section{Experiment}

In this section, we evaluate our proposed PKG framework across four distinct types of knowledge: factual, tabular, medical, and multimodal. Factual knowledge entails the model's ability to access accurate information, serving as a foundational type of knowledge crucial for numerous NLP applications (§~\ref{subsec:factual}). Tabular knowledge necessitates the model's capability to access structured knowledge in the form of tables, which is relatively scarce in the training data of LLMs (§~\ref{subsec:tabular}). Medical knowledge, being highly specialized, exhibits limited exposure within the general data (§~\ref{subsec:medical}). Lastly, multimodal knowledge poses a challenge as most LLMs are unable to process non-language information, highlighting the significance of assistance from PKG modules (§~\ref{subsec:multimodal}).

The experimental results depicted in Tables~\ref{tab:fact_table_med} and~\ref{tab:sciqa} demonstrate substantial enhancements attained through our PKG framework compared to the baseline systems. These results offer compelling evidence supporting the generalizability and effectiveness of our approach.

\subsection{Models Steup}

\paragraph{Black-Box LLMs.} We adopt one of the SOTA LLM InstructGPT3.5~\citep{instructGPT} as our target "black box" general LLMs, using the \texttt{text-davinic-002} version. With up to 175B parameters, this model is one of the largest LLMs and is pre-trained on a vast amount of internet data, which exhibits great language understanding and generation ability. However, this model can only be accessed through an API, which limits users' interaction.

\paragraph{Basic PKG Module.} Our knowledge guiding module employs the open-source and popular foundation model LLaMa-7B~\citep{llama}. It has been pre-trained on massive amounts of text data and possesses extensive world knowledge. Though its performance in many tasks may be inferior to the InstructGPTs, it can be locally fine-tuned and customized~\citep{alpaca,xu2023wizardlm,DBLP:journals/corr/abs-2304-03277,koala_blogpost_2023}, making it an effective starting point for developing a task-specific PKG module.

\paragraph{Baselines.} Our work includes three different types of baselines: (1) \textit{Direct generation without guiding}: We do not provide any background knowledge for a given task and ask the InstructGPT to generate the answer or response directly in a zero-shot manner, following the approach of prior works~\citep{gpt3,instructGPT}. (2) \textit{Generation with retrieval guiding}: We follow the retrieve-then-read paradigm~\citep{DBLP:conf/acl/ChenFWB17,DBLP:conf/naacl/YangXLLTXLL19,dpr} to retrieve related knowledge from external knowledge sources using retrieval models such as BM25~\citep{bm25} and DPR~\citep{dpr}. We fine-tune the DPR on specific tasks following the REPLUG~\citep{replug} method. InstructGPTs then generate responses based on the combination of the question and retrieved background documents. (3) \textit{Generation with self-guiding}: we adopt the InstructGPTs to generate the related background knowledge by themselves with two different methods. The first method, CoT~\cite{cot}, adopts the prompt \textit{"Let's think step-by-step"} to generate the chain-of-thought as the background knowledge. The second method, GenRead~\cite{genread}, directly requires the InstructGPTs to provide task-specific knowledge with the prompt \textit{"Please provide the background document from [domain] to [task]."}

\begin{table}
    \centering
    \caption{Evaluating on three different tasks, requiring factual (FM2), tabular (NQ-Table), and medical (MedMC-QA) knowledge. $\diamond$: we fine-tune the dense retrieval models with the task data. $\dagger$: we use InstructGPT3.5 to generate the chain-of-thoughts as the background knowledge. $\ddagger$: we use InstructGPT3.5 to generate the background documents. }
    \begin{tabular}{lcccc}
        \toprule
        \textbf{Models} & \textbf{FM2} & \textbf{NQ-Table} & \textbf{MedMC-QA}\\
        \midrule
        \midrule
        \multicolumn{4}{l}{\textit{Direct generation without guiding.}}\\
        ~~InstructGPT3.5~\citep{instructGPT} & 59.4 & 16.9 & 44.4\\
        \midrule
        \multicolumn{4}{l}{\textit{Generation with retrieval guiding.}}\\
        ~~BM25 + InstructGPT3.5~\citep{dpr} & 65.2 & 17.1 & -\\
        ~$\diamond$REPLUG + InstructGPT3.5~\citep{replug} & 65.9 & 24.3 & -\\
        \midrule
        \multicolumn{4}{l}{\textit{Generation with self-guiding.}}\\
        ~$\dagger$CoT + InstructGPT3.5~\citep{cot} & 60.4 & 21.4 & 41.5\\
        ~$\ddagger$GenRead + InstructGPT3.5~\citep{genread} & 65.5 & 23.5 & 44.4\\
        \midrule
        ~~PKG + InstructGPT3.5~(Ours) & \textbf{67.3} & \textbf{28.8} & \textbf{47.4}\\
        \bottomrule
    \end{tabular}
    \label{tab:fact_table_med}
\end{table}
\begin{table}
    \centering
    \caption{Evaluating on the ScienceQA, requiring multimodal science knowledge. $\dagger$: results from~\cite{Chameleon}. \texttt{gpt-3.5-turbo} is much more capable than \texttt{text-davinic-002}.}
    \begin{tabular}{lccccccccc}
        \toprule
        \textbf{Models} & \textbf{NAT} & \textbf{SOC} & \textbf{LAN} & \textbf{TXT} & \textbf{IMG} & \textbf{NO} & \textbf{G1-6} & \textbf{G7-12} & \textbf{Avg}\\
        \midrule
        \midrule
        \multicolumn{8}{l}{\textit{Base on gpt-3.5-turbo.}}\\
        $\dagger$ChatGPT & 78.82 & 70.98 & 83.18 & 77.37 & 67.92 & 86.13 & 80.72 & 74.03 & 78.31\\
        $\dagger$Chameleon & 81.62 & 70.64 & 84.00 & 79.77 & 70.80 & 86.62 & 81.86 & 76.53 & 79.93\\
        \midrule
        \multicolumn{8}{l}{\textit{Base on text-davinic-002.}}\\
        InstructGPT3.5 & 72.96 & 62.88 & 76.09 & 70.77 & 62.77 & 77.84 & 75.04 & 65.59 & 71.66 \\
        ~+CoT & 71.94 & 61.19 & 74.00 & 69.50 & 61.18 & 75.75 & 72.61 & 65.92 & 70.22 \\
        ~+GenRead & 72.91 & 64.68 & 76.36 & 72.14 & 63.31 & 76.66 & 74.96 & 66.91 & 72.08 \\
        ~+PKG~(Ours) & 79.35 & 82.90 & 81.91 & 79.86 & 74.32 & 83.41 & 80.80 & 80.69 & \textbf{80.76} \\
        \bottomrule
    \end{tabular}
    \label{tab:sciqa}
\end{table}

\subsection{Factual Knowledge}\label{subsec:factual}

\paragraph{Datasets and Implementation Details.} We evaluate our approach on the FM2 dataset~\citep{fm2}, which is a benchmark for fact-checking. In this task, given a factual claim, our models are required to determine whether it is true or false. We use the claim in the training set and the corresponding evidence as factual knowledge. Additionally, we sample 100k passages from English Wikipedia, each consisting of up to 256 tokens. We treat the first sentence as the input and the remaining sentences as background knowledge. Accuracy is adopted as the evaluation metric. More details can be found in Appendix A and B.

\paragraph{Results.} As shown in Table~\ref{tab:fact_table_med}, our PKG outperforms all the baseline systems for fact-checking. In comparison to direct generation, the results reveal that it is necessary to provide extra background knowledge for InstructGPTs with retrieval-based or generation-based methods. Specifically, our PKG outperforms InstructGPT3.5 by 7.9\% (67.9\% vs. 59.4\%), and outperforms REPLUG, a retrieval-based method, by 1.4\% (67.3\% vs. 65.9\%). It is noteworthy that our generation-based method does not necessitate an additional knowledge database as the retrieval-based methods. Additionally, our PKG performs better than the self-guiding method GenRead by 1.8\% (67.3\% vs. 65.5\%), indicating that our PKG can provide more useful information than the InstructGPTs themselves.

\subsection{Tabular Knowledge}\label{subsec:tabular}

\paragraph{Datasets and Implementation Details.} We evaluate the effectiveness of our approach on the NQ-Table dataset~\citep{nq_table}, which serves as a benchmark for open-domain question answering over tables. The dataset consists of questions whose answers can be found in a Wikipedia table. We adopted the question in the training set as input and the corresponding flattened table as background knowledge. Our PKG was trained to follow instructions and generate the relevant table. Exact matching is adopted as the evaluation metric. More details can be found in Appendix A and B.

\paragraph{Results.} Table~\ref{tab:fact_table_med} demonstrates the superior performance of our PKG framework over all baseline systems on the tabular knowledge-related task. Notably, our PKG outperforms InstructGPT3.5 by a substantial margin of 11.9\% (28.8\% vs. 16.9\%), and outperforms REPLUG, the retrieval-based method, by 4.5\% (28.8\% vs. 24.3\%). Furthermore, our PKG significantly outperforms the self-guiding method GenRead by 5.3\% (28.8\% vs. 23.5\%). These results demonstrate the efficacy and superiority of our approach in leveraging parametric knowledge to augment InstructGPTs for tabular knowledge-related tasks.

\subsection{Medical Knowledge}\label{subsec:medical}

\paragraph{Datasets and Implementation Details.} We evaluate the effectiveness of our approach on the MedMC-QA dataset~\citep{medqa}, which serves as a benchmark for multi-subject multi-choice medical question answering. Each question requires the use of relevant medical information as background knowledge to provide the correct answer. We use the questions in the training set as input and the corresponding medical explanation as background knowledge. Our PKG is trained to follow the instruction and generate the relevant medical background. Accuracy is the evaluation metric. Unlike the previous tasks with all Wikipedia passages as the knowledge database, we do not have access to an external medical knowledge database, and thus we do not evaluate the performance of retrieval-based methods on this task. More details can be found in Appendix A and B.

\paragraph{Results.} Our PKG framework also outperforms all baseline systems on this medical knowledge-related task, as shown in Table~\ref{tab:fact_table_med}. Specifically, our PKG outperforms InstructGPT3.5 by 3.0\% (47.4\% vs. 44.4\%). It is worth noting that the baseline self-guiding methods, CoT and GenRead, do not improve the performance of InstructGPTs. This may be due to the fact that InstructGPTs lack sufficient medical information to effectively solve this task.

\subsection{Multimodal Knowledge}\label{subsec:multimodal}

\paragraph{Datasets and Implementation Details.} Our approach is evaluated on the ScienceQA dataset~\citep{sciqa}, which presents a challenging multimodal multiple-choice question-answering task covering diverse science topics. Each question requires leveraging relevant scientific background knowledge to provide the correct answer. We use the training set's questions as input and their corresponding science lecture as background knowledge. To handle the images information, we augment our basic PKG module with the CLIP-ViT~\citep{clip} to extract visual features, which are then fused with text features using a simple one-head cross-attention mechanism in each layer of LLaMa:
\begin{equation}
    \mathcal{H} := \mathcal{H}^{txt} + \mathcal{W}^o\left(\text{softmax}\left((\mathcal{W}^q\mathcal{H}^{txt})(\mathcal{W}^k\mathcal{H}^{img})^T\right)(\mathcal{W}^v\mathcal{H}^{img})\right),
\end{equation}
where $\mathcal{W}^{o,q,k,v}$ are the linear projection, $\mathcal{H}^{txt,img}$ are the hidden states of texts and images. We adopt accuracy as the evaluation metric. More details can be found in Appendix A and B.

Similarly, this task is also difficult to obtain an external multimodal science knowledge database, retrieval-based methods are not considered. To facilitate a fair comparison of our methods, we include two additional baseline systems~\citep{Chameleon} based on the \texttt{gpt-3.5-turbo} model. The first baseline is ChatGPT direct generation, and the second is the Chameleon model, which utilizes several external tools, such as searching, OCR, and captioning. According to OpenAI, the \texttt{gpt-3.5-turbo} model is more capable than \texttt{text-davinic-002}~\citep{chatgpt_better}.

\paragraph{Results.} Table~\ref{tab:sciqa} shows that our PKG framework achieves a significant improvement in the performance of InstructGPTs on the multimodal scientific knowledge-related task. Specifically, the average accuracy is increased by 9.1\% (80.76\% vs. 71.66\%), demonstrating the effectiveness of our approach. In contrast, other guiding methods, CoT (-1.44\%) and GenRead (+0.42\%), hard to improve the performance of InstructGPTs. Moreover, our PKG framework outperforms the \texttt{gpt-3.5-turbo} based models on average by 2.45\% (80.76\% vs. 78.31\%), despite using weaker InstructGPTs.

\begin{figure}
  \centering
  \begin{subfigure}[b]{0.45\textwidth}
    \includegraphics[width=\textwidth]{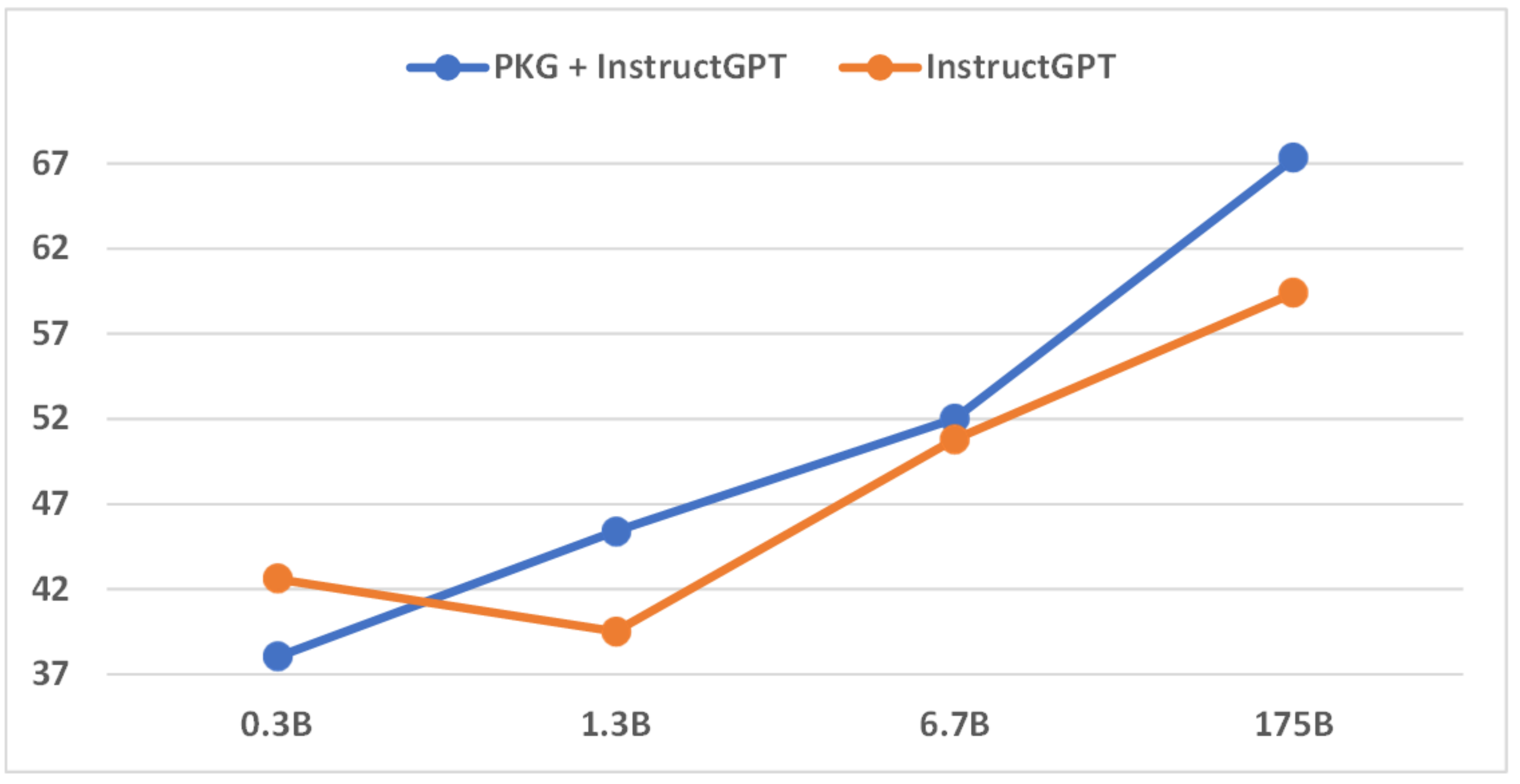}
    \caption{Accuracy on FM2.}
  \end{subfigure}
  \hfill
  \begin{subfigure}[b]{0.45\textwidth}
    \includegraphics[width=\textwidth]{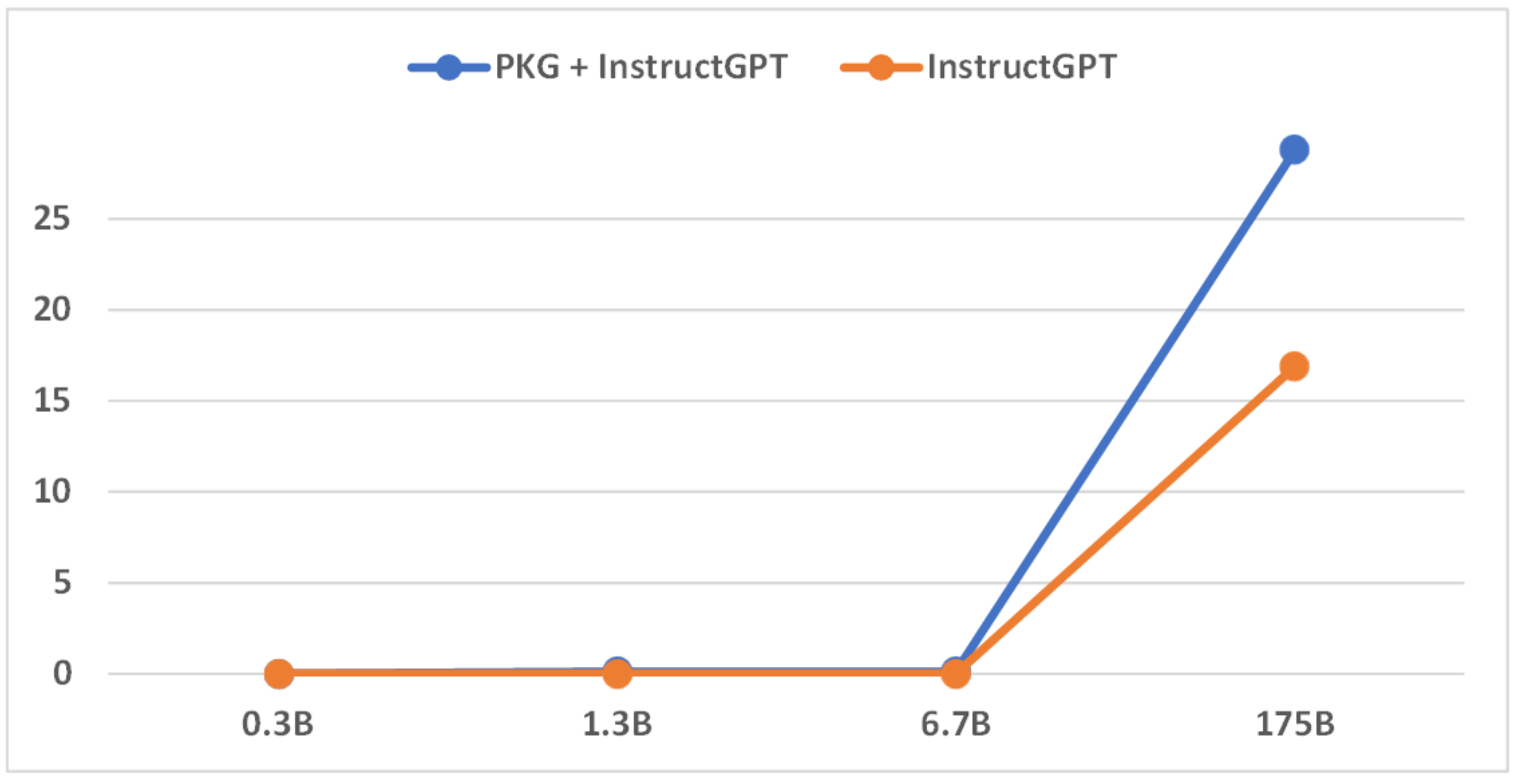}
    \caption{Exact Matching on NQ-Table.}
    \label{fig:scale_b}
  \end{subfigure}
  \hfill
  \begin{subfigure}[b]{0.45\textwidth}
    \includegraphics[width=\textwidth]{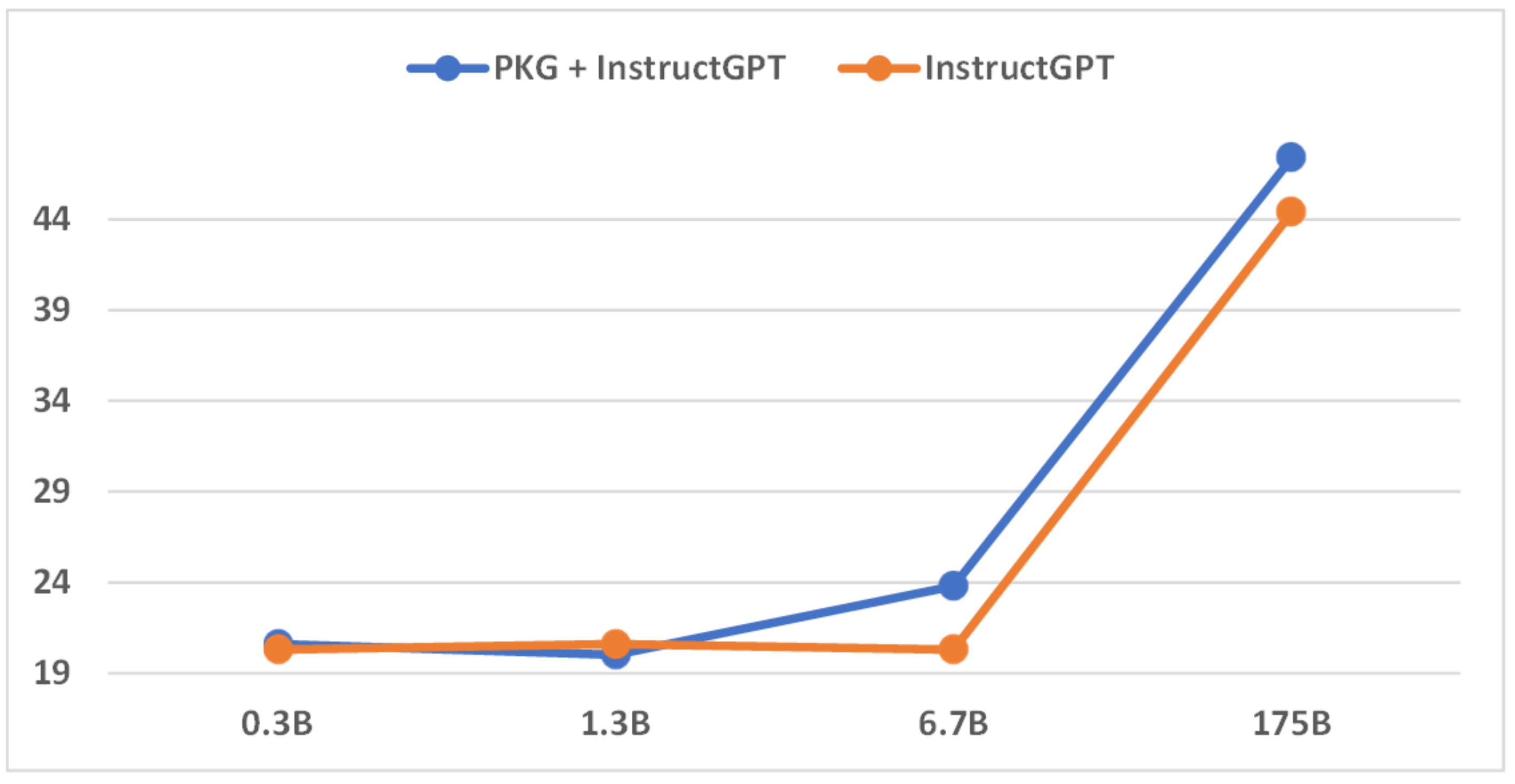}
    \caption{Accuracy on MedMC-QA.}
    \label{fig:scale_c}
  \end{subfigure}
  \hfill
  \begin{subfigure}[b]{0.45\textwidth}
    \includegraphics[width=\textwidth]{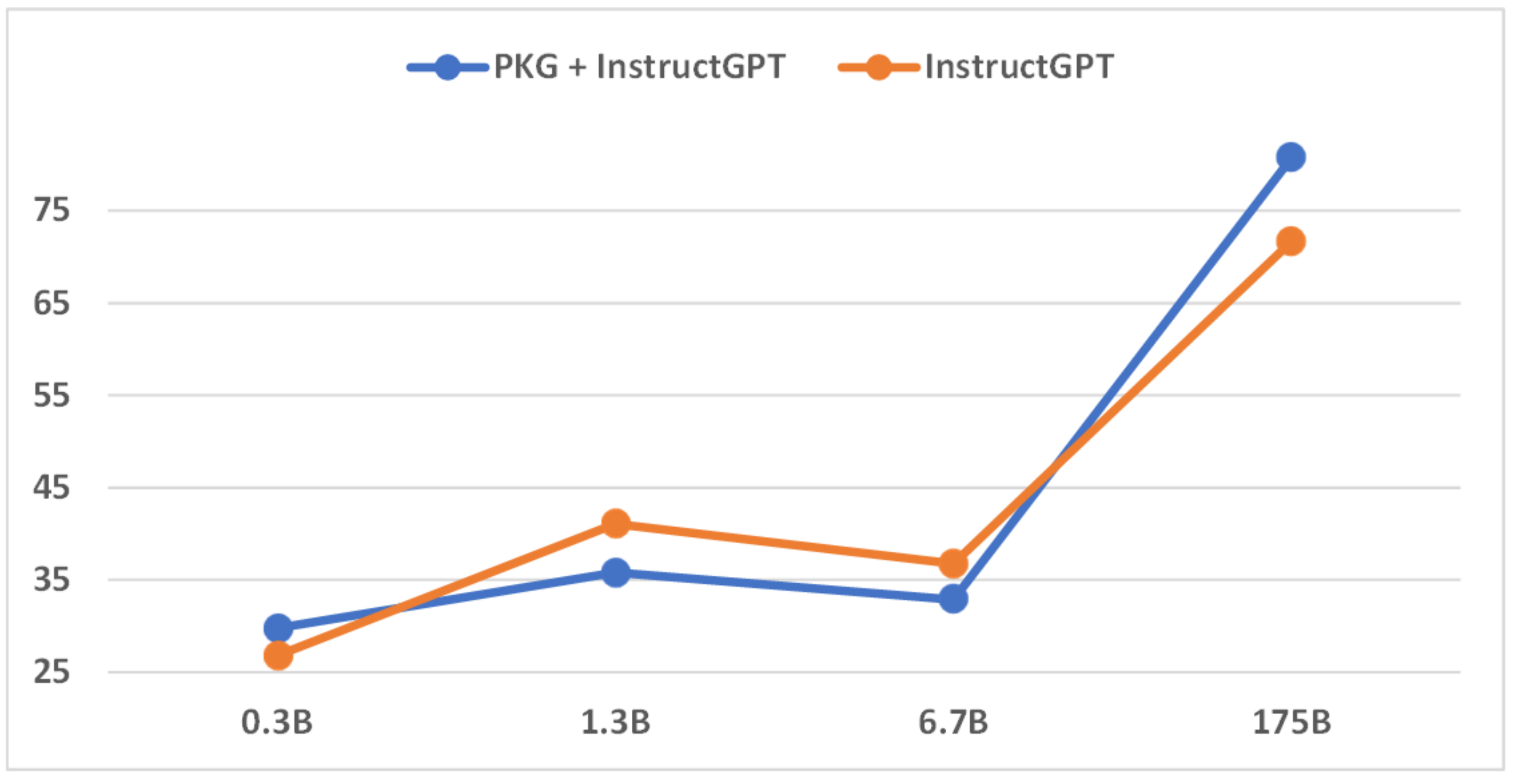}
    \caption{Average Accuracy on ScienceQA}
  \end{subfigure}
  \caption{Comparing our PKGs framework with the direct generation on various types of InstructGPT. The number indicates the number of parameters in the InstructGPT. 0.3B: text-ada-001, 1.3B: text-babbage-001, 6.7B: text-curie-001, 175B: text-davinci-002.}
  \label{fig:scale}
\end{figure}

\begin{table}
    \centering
    \caption{Comparing various sizes of language models as the basic PKG modules.}
    \begin{tabular}{lcccc}
        \toprule
        \textbf{Basic PKG} & \textbf{FM2} & \textbf{NQ-Table} & \textbf{MedMC-QA} & \textbf{ScienceQA}\\
        \midrule
        \midrule
        LLaMa-7B~\citep{llama} & \textbf{67.3} & \textbf{28.8} & \textbf{47.4} & \textbf{80.8}\\
        \midrule
        OPT-2.7B~\citep{opt} & 59.6 & 17.9 & 34.4 & 79.5\\
        OPT-1.3B & 58.2 & 16.5 & 33.9 & 77.0\\
        OPT-0.3B & 56.4 & 14.6 & 31.7 & 68.7\\
        \bottomrule
    \end{tabular}
    \label{tab:basic_pkg}
\end{table}

\subsection{Analysis}

\paragraph{Scale of LLMs.} Figure~\ref{fig:scale} presents the impact of our PKG framework on several "black-box" LMs, including \texttt{text-ada-001}, \texttt{text-babbage-001}, \texttt{text-curie-001}, and \texttt{text-davinci-002}. The results suggest that the effectiveness of our approach is correlated with the size of the LMs, with larger LMs benefiting more from our PKGs than smaller ones. Specifically, in Figure~\ref{fig:scale_b}, the small LMs show negligible exact matching scores on the tabular task, with or without the background knowledge from our PKGs, while the LLMs exhibit significantly better performance. In Figure~\ref{fig:scale_c}, the 0.3B and 1.3B LMs perform similarly on the medical domain task, while the 6.7B LM shows improved performance with the additional knowledge. This difference can be attributed to the relatively weaker language understanding capabilities of smaller LMs, which struggle to reason over contexts and generate the correct responses even with relevant knowledge from our PKGs. These observations align with the emergent abilities of LLMs, as discussed in~\cite{DBLP:journals/corr/abs-2206-07682}. Therefore, the scale of LLMs is a critical factor for achieving better performance.

\paragraph{Scale of PKGs.} We conducted an investigation of various sizes of language models as basic PKG modules in Table~\ref{tab:basic_pkg}. Since LLaMa-7B is the smallest model in the LLaMa family, we conducted experiments on the OPT family~\citep{opt}, another open-source large-scale language model with a similar structure to LLaMa. Our observations reveal that larger basic PKGs tend to exhibit superior performance. For example, increasing the number of parameters from 1.3B to 2.7B leads to performance improvements of 1.4\% on FM2, 1.4\% on NQ-Table, 0.5\% on MedMC-QA, and 2.5\% on ScienceQA, which is consistent with the scaling law~\citep{DBLP:journals/corr/abs-2001-08361}.

\begin{table}
    \centering
    \caption{Examples of background documents generated by our PKGs to guide different tasks. Clues to answer the input are highlighted in \textcolor{blue}{blue} within the documents.}
    \begin{tabular}{p{2.9cm}|p{10cm}}
        \toprule
        Input (FM2) & The Natural is a book about Roy Hobbs a natural southpaw boxer who goes on to win the heavyweight title from Boom Boom Mancini.\\
        \midrule
        PKG Guiding & This is \textcolor{blue}{a story about a baseball player, Roy Hobbs}, who is shot by a woman, Harriet Bird, and then recovers to become a great player.\\
        \bottomrule
        \toprule
        Input (NQ-Table) & Batman The Enemy Within episode 5 release date\\
        \midrule
        PKG Guiding & \begin{tabular}{|c|c|c|c|}
        \multicolumn{4}{c}{Batman: The Enemy Within}\\
        \hline
        Ep & Title &  Directed by & Release date \\
        \hline
        1 & "The Enigma" & Kent Mudle & August 8, 2017\\
        2 & "The Pact" & Matthew Leach & October 3, 2017\\
        3 & "Fractured Mask" & Sean Manning & November 21, 2017\\
        4 & "What Ails You" & Chris Rieser & January 23, 2018\\
        5 & "Same Stitch" & Kent Mudle & \textcolor{blue}{March 27, 2018}\\
        \hline
      \end{tabular}
        \\
        \bottomrule
        \toprule
        Input (MedMC-QA) & Calcium ions triggers muscle contraction by binding to: Options: (A) Actin (B) Myosin (C) Troponin (D) Tropomyosin\\
        \midrule
        PKG Guiding & Troponin is a complex of three proteins that are attached to actin. \textcolor{blue}{Calcium ions bind to troponin-C}, which then causes the troponin-I and troponin-T to shift and allow the tropomyosin to move off of the active site of the myosin so that the myosin head can then bind to actin and cause contraction.\\
        \bottomrule
        \toprule
        Input (ScienceQA) \includegraphics[width=1.5cm]{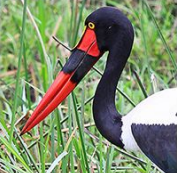} & Which animal's neck is also adapted for hunting prey while keeping the rest of its body still? Context: Saddle-billed storks live near wetlands and lakes. They eat mostly fish. The 's neck helps it grab fish while keeping the rest of its body still. If the stork had to move its body, it might scare the fish away. Figure: saddle-billed stork. A black and white bird is standing in the grass. Options: (A) northern pintail (B) black-headed heron\\
        \midrule
        PKG Guiding & Look at the picture of the saddle-billed stork. The saddle-billed stork has a long neck. Its neck is adapted for hunting prey while keeping the rest of its body still. Now look at each animal. Figure out which animal has a similar adaptation. \textcolor{blue}{The black-headed heron has a long neck. Its neck is adapted for hunting prey while keeping the rest of its body still.} The northern pintail has a short neck. Its neck is not adapted for hunting prey while keeping the rest of its body still.\\
        \bottomrule
    \end{tabular}
    \label{tab:pkg_example}
\end{table}

\paragraph{Examples of Generated Background Documents.} Table~\ref{tab:pkg_example} presents examples of background documents generated by our PKGs to assist LLMs in different tasks. For the factual task, our PKG can supply input-related factual information to support or refute the input, such as the example of Roy Hobbs being a baseball player and not a boxer. For the tabular task, our PKG can offer an input-related background table, like the episode table of Batman. For the medical task, our PKG can provide relevant medical knowledge, such as the background of calcium ions. For the multimodal task, our PKG can produce a document based on text information while taking into account the image context in the input, for example, noting that the bird in the image has a long neck. Additional examples can be found in Appendix D.

\section{Conclusion}

In this work, we propose the novel \textbf{Parametric Knowledge Guiding (PKG)} framework to enhance the performance of "black-box" LLMs on domain-specific tasks by equipping them with a knowledge-guiding module. Our approach allows for access to relevant knowledge at runtime without altering the LLM's parameters. The experiments demonstrate the effectiveness of our PKG framework for various domain knowledge-intensive tasks.

\textbf{Limitation and Future Work.} Although our PKGs have shown strong performance on the presented datasets, they may still suffer from hallucination errors, leading to the provision of incorrect background knowledge. We provide examples of such errors in Appendix E. Combining our approach with retrieval methods to enhance generative faithfulness is a promising direction for future research.

\bibliographystyle{elsarticle-harv}
\bibliography{nips}

\appendix

\section{Datasets and Splits}

- Fool Me Twice (FM2)~\citep{fm2} contains a set of claims with evidence that were originally scarped from Wikipedia.

- Natural Questions Over Tables (NQ-Table)~\citep{nq_table} were mined from real Google search queries and the answers are spans in Wikipedia tables identified by human annotators.

- Multi-Subject Multi-Choice Dataset for Medical domain (MedMC-QA)~\citep{medqa} contains a set of real-world medical entrance exam questions and answers.

- Multimodal Reasoning for Science Question Answering (ScienceQA)~\citep{sciqa} consists of multimodal multiple-choice questions with a diverse set of science topics.

Table~\ref{tab:dataset} shows the dataset splits and statistics.

\begin{table}[h!]
    \centering
    \caption{Datasets splits and statistics. For MedMC-QA, labels in the test are hidden, so the model performance is evaluated on the validation set.}
    \begin{tabular}{l|ccccc}
        \toprule
        Datasets & Domain & Train & Valid & Test & Test labels\\
        \midrule
        FM2~\citep{fm2} & Factual & 10,419 & 1,169 & 1,380 & Public\\
        NQ-Table~\citep{nq_table} & Tabular & 9,594 & 1,068 & 959 & Public\\
        MedMC-QA~\citep{medqa} & Medical & 160,869 & 4,183 & 6,150 & Private\\
        ScienceQA~\citep{sciqa} & Multimodal & 12,726 & 4,241 & 4,241 & Public\\
        \bottomrule
    \end{tabular}
    \label{tab:dataset}
\end{table}

\section{Implementation Details}

We use LLaMa-7B~\citep{llama} as our backbone models to implement the PKG modules. We use AdamW as the optimizer, with 10\% warmup steps. We use 8 V100 GPUs for training PKG modules. The open-source code \textit{LLaMa-X}\footnote{\url{https://github.com/AetherCortex/Llama-X}} is widely used in our experiments. We refer to more individual implementation details in Table~\ref{tab:Hyper}.

\begin{table}
    \centering
    \caption{Hyperparameters settings of our PKG modules on different tasks.}
    \begin{tabular}{l|cccc}
        \toprule
        Settings & FM2 & NQ-Table & MedMC-QA & ScienceQA\\
        \midrule
        Peak learning rate & 2e-5 & 2e-5 & 2e-5 & 2e-5\\
        Total batch size & 64 & 32 & 32 & 32\\
        Total training epochs & 3 & 10 & 3 & 5\\
        Warmup Schedule & cosine & cosine & cosine & cosine\\
        Warmup ratio & 0.1 & 0.1 & 0.1 & 0.1\\
        \bottomrule
    \end{tabular}
    \label{tab:Hyper}
\end{table}

We implement other baseline methods based on the following repositories:

- BM25 + GPT3.5: \url{https://github.com/castorini/pyserini}

- REPLUG + GPT3.5: \url{https://github.com/facebookresearch/DPR/tree/main}

- CoT + GPT3.5: \url{https://github.com/kojima-takeshi188/zero_shot_cot}

- GenRead + GPT3.5: \url{https://github.com/wyu97/GenRead}

\section{All Experiment Results of Figure 3}

In Figure 3, we compare our PKGs framework with the direct generation on various types of LMs. We include all results in Table~\ref{tab:results}.

\begin{table}
    \centering
    \caption{All experiments results of Figure 3 for different sizes of LMs.}
    \begin{tabular}{lcccc}
        \toprule
        Methods & FM2 & NQ-Table & MedMC-QA & ScienceQA\\
        \midrule
        PKG-Davinci & \textbf{67.3} & \textbf{28.8} & \textbf{47.4} & \textbf{80.76}\\
        PKG-Curie & 52.0 & 0.1 & 23.8 & 32.87\\
        PKG-Babbage & 45.4 & 0.1 & 20.0 & 35.77\\
        PKG-Ada & 38.0 & 0.0 & 20.6 & 29.76\\
        \midrule
        Direct-Davinci & 59.4 & 16.9 & 44.4 & 71.66\\
        Direct-Curie & 50.8 & 0.0 & 20.3 & 36.76\\
        Direct-Babbage & 39.5 & 0.0 & 20.6 & 41.08\\
        Direct-Ada & 42.6 & 0.0 & 20.3 & 26.79\\
        \bottomrule
    \end{tabular}
    \label{tab:results}
\end{table}

\section{Case Studies}

Table~\ref{tab:pkg_example_fm2}~\ref{tab:pkg_example_nq_table}~\ref{tab:pkg_example_medqa}~\ref{tab:pkg_example_sciqa} presents more examples of background documents generated by our baseline methods (CoT and GenRead) and PKGs for different tasks. We can notice that our PKGs can provide clues for LLMs to answer specific questions. Table~\ref{tab:pkg_example_nq_table_retrieval} also compares our PKGs with the retrieval-based methods. We can find that the retrieval methods cannot provide relevant background documents to answer the question.

\begin{table}
    \centering
    \caption{Examples of hallucination errorsbackground documents generated by our baseline methods and PKGs for FM2. Clues to answering the input are highlighted in \textcolor{blue}{blue} within the documents. The information that will disrupt LLMs is highlighted in \textcolor{red}{red}.}
    \begin{tabular}{p{2.9cm}|p{10cm}}
        \toprule
        Input (FM2) & The Natural is a book about Roy Hobbs a natural southpaw boxer who goes on to win the heavyweight title from Boom Boom Mancini.\\
        \midrule
        CoT Guiding & The first part of the claim is that The Natural is a book. This is true - The Natural is a novel written by Bernard Malamud. The second part of the claim is that \textcolor{red}{the book is about Roy Hobbs, a natural southpaw boxer. This is also true} - the novel follows the character of Roy Hobbs, a talented young baseball player who is drafted by a major league team. The third part of the claim is that Hobbs goes on to win the heavyweight title from Boom Boom Mancini. This is false - in the novel, Hobbs does not win the heavyweight title from Boom Boom Mancini.\\
        \midrule
        GenRead Guiding & \textcolor{blue}{The Natural is a 1984 American sports drama film based on Bernard Malamud's 1952 baseball novel of the same name}, directed by Barry Levinson and starring Robert Redford, Glenn Close, and Robert Duvall. The film, like the book, recounts the experiences of Roy Hobbs, an extraordinarily talented left-handed pitcher, who becomes a late-blooming professional baseball player with the New York Knights.\\
        \midrule
        PKG Guiding & This is \textcolor{blue}{a story about a baseball player, Roy Hobbs}, who is shot by a woman, Harriet Bird, and then recovers to become a great player.\\
        \bottomrule
        \toprule
        Input (FM2) & Goodbye, Mr. Chips' story happens in Canada.\\
        \midrule
        CoT Guiding & First, let's recall what the story is about. \textcolor{blue}{The story is about a schoolteacher, Mr. Chips, who spends his entire life teaching at a boys' boarding school in England}. Now, let's think about whether the story could have taken place in Canada. Could a schoolteacher in Canada spend his entire life teaching at a boys' boarding school? It's possible, but it seems unlikely. Could a schoolteacher in Canada spend his entire life teaching at any kind of school? Yes, that's possible. Therefore, the claim that the story happens in Canada is false.\\
        \midrule
        GenRead Guiding & Goodbye, Mr. Chips is a novel by James Hilton, first published in 1934. The novel tells the story of a schoolteacher, Mr. Chipping, who is nicknamed "Chips". The novel was Hilton's best-selling book of all time, and was adapted into two films and a television series. The novel is set in the fictional public school of Brookfield, in the English county of Leicestershire. \textcolor{red}{However, in the film adaptations, the school is moved to the Canadian Rockies}.\\
        \midrule
        PKG Guiding & Goodbye, Mr. Chips is a 1934 British drama film based on the 1934 novel of the same name by James Hilton, which tells \textcolor{blue}{the story of a school headmaster at an English boys' school} from 1870 to 1935.\\
        \bottomrule
    \end{tabular}
    \label{tab:pkg_example_fm2}
\end{table}
\begin{table}
    \centering
    \caption{Examples of background documents generated by our baseline methods and PKGs for NQ-Table. Clues to answering the input are highlighted in \textcolor{blue}{blue} within the documents. The information that will disrupt LLMs is highlighted in \textcolor{red}{red}.}
    \begin{tabular}{p{2.9cm}|p{10cm}}
        \toprule
        Input (NQ-Table) & Batman The Enemy Within episode 5 release date\\
        \midrule
        CoT Guiding & 1. When is episode 5 of Batman the enemy released? 2. What is the release date of episode 5 of Batman the enemy? \textcolor{red}{The release date of episode 5 of Batman the enemy is October 10, 2016}.\\
        \midrule
        GenRead Guiding & "Batman: The Enemy Within" is an episodic point-and-click graphic adventure video game developed and published by Telltale Games and distributed by Warner Bros. Interactive Entertainment under its DC Entertainment label. The game is a sequel to 2016's "Batman: The Telltale Series", and was released for Microsoft Windows, macOS, PlayStation 4, Xbox One, Android, and iOS. The first episode was released on August 8, 2017, with the fifth and \textcolor{blue}{final episode released on March 27, 2018}.\\
        \midrule
        PKG Guiding & \begin{tabular}{|c|c|c|c|}
        \multicolumn{4}{c}{Batman: The Enemy Within}\\
        \hline
        Ep & Title &  Directed by & Release date \\
        \hline
        1 & "The Enigma" & Kent Mudle & August 8, 2017\\
        2 & "The Pact" & Matthew Leach & October 3, 2017\\
        3 & "Fractured Mask" & Sean Manning & November 21, 2017\\
        4 & "What Ails You" & Chris Rieser & January 23, 2018\\
        5 & "Same Stitch" & Kent Mudle & \textcolor{blue}{March 27, 2018}\\
        \hline
      \end{tabular}
        \\
        \bottomrule
        \toprule
        Input (NQ-Table) & the old man and the sea page count\\
        \midrule
        CoT Guiding & \textcolor{red}{The old man is fishing and he catches a big fish. He struggles to bring the fish in, and finally does. Now, he has to get the fish home. So, he ties it to his boat and starts rowing. The old man is rowing for a long time and finally makes it back to shore. He drags the fish up onto the beach and starts to cut it open. Now, we're at the part where the old man is cutting open the fish. He's cut open the belly and is starting to pull out the guts. He's got a knife in one hand and is pulling out the guts with the other. The old man is still cutting open the fish. He's almost done now. He's got the knife in one hand and the guts in the other. He's about to pull out the last of the guts when he suddenly feels something tugging on his line. The old man has been pulled out to sea by the big fish! He's struggling to get back to shore, but the fish is too strong. The old man is fighting for his life now. The old man is still struggling to get back to shore. He's getting tired now, and the fish is still pulling him out to sea. The old man knows he's going to die soon. The old man finally dies and is pulled out to sea by the big fish}.\\
        \midrule
        GenRead Guiding & \textcolor{red}{The Old Man and the Sea is a novel by Ernest Hemingway that was first published in 1952. The novel tells the story of an aging Cuban fisherman who struggles to catch a giant marlin. The Old Man and the Sea was published in 1952 in Life magazine. The novel was adapted into a 1958 film of the same name, starring Spencer Tracy. The Old Man and the Sea won the Pulitzer Prize for Fiction in 1953}.\\
        \midrule
        PKG Guiding & \begin{tabular}{|c|c|c|c|}
        \multicolumn{3}{c}{The Old Man and the Sea}\\
        \hline
        Author & Language & Genre & Pages\\
        \hline
        Ernest Hemingway & English & Literary Fiction & \textcolor{blue}{127}\\
        \hline
      \end{tabular}
        \\
        \bottomrule
    \end{tabular}
    \label{tab:pkg_example_nq_table}
\end{table}
\begin{table}
    \centering
    \caption{Examples of background documents generated by our baseline methods and PKGs for MedMC-QA. Clues to answering the input are highlighted in \textcolor{blue}{blue} within the documents. The information that will disrupt LLMs is highlighted in \textcolor{red}{red}.}
    \begin{tabular}{p{2.9cm}|p{10cm}}
        \toprule
        Input (MedMC-QA) & Calcium ions triggers muscle contraction by binding to: Options: (A) Actin (B) Myosin (C) Troponin (D) Tropomyosin (Correct Answer: D)\\
        \midrule
        CoT Guiding & First, we need to understand what a muscle contraction is. A muscle contraction is when the muscle fibers in the muscle shorten. This is caused by the sliding of the actin and myosin filaments past each other. Next, we need to understand what role calcium ions play in muscle contraction. Calcium ions are needed for the muscle contraction to occur because they are needed to trigger the release of the myosin head from the actin filament. \textcolor{blue}{So, the answer is (C) Troponin}.\\
        \midrule
        GenRead Guiding & Calcium ions play an important role in muscle contraction. \textcolor{blue}{When calcium ions bind to troponin}, they cause a conformational change in the protein that allows myosin and actin to interact. This interaction between myosin and actin is what causes muscle contraction.\\
        \midrule
        PKG Guiding & Troponin is a complex of three proteins that are attached to actin. \textcolor{blue}{Calcium ions bind to troponin-C}, which then causes the troponin-I and troponin-T to shift and allow the tropomyosin to move off of the active site of the myosin so that the myosin head can then bind to actin and cause contraction.\\
        \bottomrule
        \toprule
        Input (MedMC-QA) & Which of the following statements is true regarding medical aboion? Options: (A) Ultrasound should be done in all cases (B) If the patient has an IUCD in-situ, it doesn't need to be removed (C) Can only be done up to 72 days (D) Only a person ceified under MTP act can perform medical termination of pregnancy (Correct Answer: D)\\
        \midrule
        CoT Guiding & First, we need to understand what medical abortion is. Medical abortion is the use of medication to end a pregnancy. Next, we need to consider the options given. Option A is not necessarily true - ultrasound may not be needed in all cases, but it can be helpful in determining how far along the pregnancy is. Option B is also not necessarily true - in some cases, the IUCD (intrauterine contraceptive device) may need to be removed before the medical abortion can be performed. \textcolor{red}{Option C is generally true - medical abortions can usually be performed up to 72 days after the last menstrual period}. \textcolor{blue}{Option D is also generally true - only a certified healthcare provider can perform a medical abortion}.\\
        \midrule
        GenRead Guiding & \textcolor{red}{Medical abortion is a safe and effective way to end a pregnancy. It can be done up to 72 days after the first day of the last menstrual period. A medical abortion can be done with pills or with a shot}.\\
        \midrule
        PKG Guiding & Medical aboion can be done up to 9 weeks and 6 days. Ultrasound is not required for medical aboion. If the patient has an IUCD in-situ, it should be removed. \textcolor{blue}{Only a person ceified under MTP act can perform medical termination of pregnancy}.\\
        \bottomrule
    \end{tabular}
    \label{tab:pkg_example_medqa}
\end{table}
\begin{table}
    \centering
    \caption{Examples of background documents generated by our baseline methods and PKGs for ScienceQA. Clues to answering the input are highlighted in \textcolor{blue}{blue} within the documents. The information that will disrupt LLMs is highlighted in \textcolor{red}{red}.}
    \begin{tabular}{p{2.9cm}|p{10cm}}
        \toprule
        Input (ScienceQA) \includegraphics[width=1.5cm]{bird.png} & Which animal's neck is also adapted for hunting prey while keeping the rest of its body still? Context: Saddle-billed storks live near wetlands and lakes. They eat mostly fish. The 's neck helps it grab fish while keeping the rest of its body still. If the stork had to move its body, it might scare the fish away. Figure: saddle-billed stork. A black and white bird is standing in the grass. Options: (A) northern pintail (B) black-headed heron (Correct Answer: B)\\
        \midrule
        CoT Guiding & Option A is incorrect because northern pintails are ducks and do not have necks adapted for hunting prey. \textcolor{blue}{Option B is correct because black-headed herons are wading birds that use their long necks to help them catch fish}.\\
        \midrule
        GenRead Guiding & \textcolor{blue}{The answer to the question is option (B), the painted stork}. The painted stork's neck is adapted for hunting prey while keeping the rest of its body still, which helps it grab fish while minimizing movement that could scare the fish away.\\
        \midrule
        PKG Guiding & Look at the picture of the saddle-billed stork. The saddle-billed stork has a long neck. Its neck is adapted for hunting prey while keeping the rest of its body still. Now look at each animal. Figure out which animal has a similar adaptation. \textcolor{blue}{The black-headed heron has a long neck. Its neck is adapted for hunting prey while keeping the rest of its body still.} The northern pintail has a short neck. Its neck is not adapted for hunting prey while keeping the rest of its body still.\\
        \bottomrule
        \toprule
        Input (ScienceQA) \includegraphics[width=2.5cm]{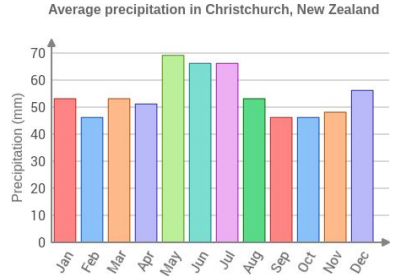} & Which month is the wettest on average in Christchurch? Context: Use the graph to answer the question below. A series of photographs showing a variety of colorful objects. Options: (A) April (B) December (C) May (Correct Answer: C)\\
        \midrule
        CoT Guiding & \textcolor{red}{The wettest month on average in Christchurch is December}.\\
        \midrule
        GenRead Guiding & \textcolor{red}{The graph shows that December is the wettest month on average in Christchurch}.\\
        \midrule
        PKG Guiding & Scientists record data from places around the world. Precipitation, or rain and snow, is one type of climate data. A bar graph can be used to show the average amount of precipitation each month. Months with taller bars have more precipitation on average. To describe the average precipitation trends in Christchurch, look at the graph. Choice "Apr" is incorrect. \textcolor{red}{Choice "May" is incorrect}. Choice "Dec" is incorrect. \textcolor{blue}{May has an average monthly precipitation of about 70 millimeters. This is higher than in any other month. So, May is the wettest month on average}.\\
        \bottomrule
    \end{tabular}
    \label{tab:pkg_example_sciqa}
\end{table}

\begin{table}
    \centering
    \caption{Examples of background documents generated by our baseline methods and PKGs for NQ-Table. Clues to answering the input are highlighted in \textcolor{blue}{blue} within the documents.}
    \begin{tabular}{p{2.9cm}|p{10cm}}
        \toprule
        Input (NQ-Table) & Batman The Enemy Within episode 5 release date\\
        \midrule
        BM25 Retrieved & is either visited by Bruce or decides to become the hero's archenemy. However, if he was a criminal, he is shown playing with a doll version of Bruce, which he promises to see again. All episodes below were released for Windows, macOS, PlayStation, Xbox One, and mobile platforms on the dates given. The Nintendo Switch version was released as a single package on October 2, 2018. "Batman: The Enemy Within" was considered to be an improvement over its predecessor, earning praise for its story, choices, action sequences, and portrayal of the Batman mythos. However, the presence of technical issues, and\\
        \midrule
        REPLUG Retrieved & Babylon 5: The Legend of the Rangers Babylon 5: The Legend of the Rangers (subtitled: To Live and Die in Starlight) is the fifth telefilm set in the "Babylon 5" universe (not including the pilot,). Originally airing January 19, 2002 on the Sci Fi Channel, it was written by J. Michael Straczynski and directed by Mike Vejar. Though shot as a pilot for a possible new series, it aired opposite NFL playoffs and the subsequent poor ratings led to it not being picked up. As the Shadow War ended, hundreds of civilizations were devastated. It is up to the\\
        \midrule
        PKG Guiding & \begin{tabular}{|c|c|c|c|}
        \multicolumn{4}{c}{Batman: The Enemy Within}\\
        \hline
        Ep & Title &  Directed by & Release date \\
        \hline
        1 & "The Enigma" & Kent Mudle & August 8, 2017\\
        2 & "The Pact" & Matthew Leach & October 3, 2017\\
        3 & "Fractured Mask" & Sean Manning & November 21, 2017\\
        4 & "What Ails You" & Chris Rieser & January 23, 2018\\
        5 & "Same Stitch" & Kent Mudle & \textcolor{blue}{March 27, 2018}\\
        \hline
      \end{tabular}
        \\
        \bottomrule
        \toprule
        Input (NQ-Table) & the old man and the sea page count\\
        \midrule
        BM25 Retrieved & by Magneto's Sentinels for plotting against Magneto. In "JLA/Avengers", Count Nefaria is seen in \#4 among the other villains enthralled by Krona to defend his stronghold. He is shown fighting Superman in a panel spreading across two-pages. In the pages of "Old Man Logan", the elderly Logan awoke on Earth-616 and had a flashback to where Count Nefaria, Red Skull, Baron Blood, Spiral, and Whirlwind were standing over the dead bodies of the superheroes the day when the villains rose and the heroes fell. Count Nefaria appears in the "What If?" story "What If the X-Men Died on their First\\
        \midrule
        REPLUG Retrieved & The Old Man and the Sea The Old Man and the Sea is a short novel written by the American author Ernest Hemingway in 1951 in Cuba, and published in 1952. It was the last major work of fiction by Hemingway that was published during his lifetime. One of his most famous works, it tells the story of Santiago, an aging Cuban fisherman who struggles with a giant marlin far out in the Gulf Stream off the coast of Cuba. In 1953, "The Old Man and the Sea" was awarded the Pulitzer Prize for Fiction, and it was cited by\\
        \midrule
        PKG Guiding & \begin{tabular}{|c|c|c|c|}
        \multicolumn{3}{c}{The Old Man and the Sea}\\
        \hline
        Author & Language & Genre & Pages\\
        \hline
        Ernest Hemingway & English & Literary Fiction & \textcolor{blue}{127}\\
        \hline
      \end{tabular}
        \\
        \bottomrule
    \end{tabular}
    \label{tab:pkg_example_nq_table_retrieval}
\end{table}

\section{Errors}

\begin{table}[h!]
    \centering
    \caption{Examples of hallucination errors. \textcolor{red}{red}: indicates the errors.}
    \begin{tabular}{p{2.9cm}|p{10cm}}
        \toprule
        Input & who won game 4 of the 2000 nba finals (Correct Answer: Los Angeles Lakers)\\
        \midrule
        PKG Guiding & \begin{tabular}{|c|c|c|c|}
        \multicolumn{4}{c}{2000 NBA Finals}\\
        \hline
        Game & Home Team & Result & Road Team \\
        \hline
        Game 1 & Los Angeles Lakers & 116-86 & Portland Trail Blazers\\
        Game 2 & Los Angeles Lakers & 100-86 & Portland Trail Blazers\\
        Game 3 & Portland Trail Blazers & 86-80 & Los Angeles Lakers\\
        Game 4 & Portland Trail Blazers & \textcolor{red}{89-78} & Los Angeles Lakers\\
        \hline
      \end{tabular}
        \\
        \bottomrule
    \end{tabular}
    \label{tab:pkg_example_hall}
\end{table}

Table~\ref{tab:pkg_example_hall} presents a hallucination error of our PKGs.

\newpage

\section{Instruction Formats}

- FM2:

\texttt{\scriptsize
Below is an instruction that describes a task, paired with an input that provides further context.\\
Write a response that appropriately completes the request.\\
\#\#\# Instruction:\\
Generate a background document from Wikipedia to support or refute the statement.\\
\#\#\# Input:\\
Statement: xxx\\
\#\#\# Response:\\
<background fact>\\
}

- NQ-Table:

\texttt{\scriptsize
Below is an instruction that describes a task, paired with an input that provides further context.\\
Write a response that appropriately completes the request.\\
\#\#\# Instruction:\\
Generate a background table from Wikipedia to answer the given question.\\
\#\#\# Input:\\
Question: xxx\\
\#\#\# Response:\\
<background table>\\
}

-MedMC-QA

\texttt{\scriptsize
Below is an instruction that describes a task, paired with an input that provides further context.\\
Write a response that appropriately completes the request.\\
\#\#\# Instruction:\\
Generate a background document from the medical domain to answer the given question.\\
\#\#\# Input:\\
Question: xxx\\
\#\#\# Response:\\
<background medical knowledge>\\
}

-ScienceQA: We follow the "QCM-LE" format in MM-CoT~\citep{DBLP:journals/corr/abs-2302-00923}, where "Q" is the question, "C" is the context, "M" is the choices, "L" is the lecture and "E" is the explanation. Please refer to the paper of MM-CoT for more details.

\section{Prompt}

- FM2: "{background} \textbackslash{n}\textbackslash{n} claim: {query} \textbackslash{n}\textbackslash{n} Is the claim true or false?"

- NQ-Table: "Refer to the background below and answer the following question with just a few words. The answer should be less than 5 words.\textbackslash{n}\textbackslash{n} Background: {background}\textbackslash{n}\textbackslash{n} Question: {question}\textbackslash{n}\textbackslash{n} Answer:"

- MedMC-QA: "Refer to the medical background below and answer the following question.\textbackslash{n} Background: {background}\textbackslash{n}\textbackslash{n}Question: {question}\textbackslash{n}Options: {options}\textbackslash{n}\textbackslash{n}Please only choose the answer from options. The answer is:"

- ScienceQA: "Question: {question}\textbackslash{n}BECAUSE: {background}\textbackslash{n}Options: {options}\textbackslash{n}Please only choose the answer from options. The answer is:"

\end{document}